\documentclass[10pt,twocolumn,letterpaper]{article}

\usepackage{wacv}
\usepackage{times}
\usepackage{epsfig}
\usepackage{graphicx}
\usepackage{amsmath}
\usepackage{amssymb}
\usepackage{booktabs}
\usepackage{color}
\usepackage{multirow}
\usepackage{soul}
\usepackage{xcolor}
\usepackage{url}
\usepackage{cite}
\usepackage{floatrow}
\usepackage[draft,inline,nomargin]{fixme}
\fxsetup{
    status=draft,
    author=,
    layout=inline,
    theme=color
}

\wacvfinalcopy 

\def\wacvPaperID{****} 

\newcommand{\ra}[1]{\renewcommand{\arraystretch}{#1}}
\newcommand{\xhdr}[1]{\vspace{6pt} \noindent {\textbf{#1} }}

\ifwacvfinal\fi
\begin{document}

\title{Fully-Coupled Two-Stream Spatiotemporal Networks \\
for Extremely Low Resolution Action Recognition}

\author{Mingze Xu$^{1}$\thanks{Part of this work was done when Mingze Xu and
    Aidean Sharghi were interns at Midea Corporate Research Center, San Jose, CA.}
    \hspace{0.5cm} Aidean Sharghi$^{2}$\footnotemark[1]
    \hspace{0.5cm} Xin Chen$^{3}$\thanks{Corresponding author.}
    \hspace{0.5cm} David J. Crandall$^{1}$ \\
    $^{1}$Indiana University, Bloomington, IN \\
    $^{2}$University of Central Florida, Orlando, FL \\
    $^{3}$Midea Corporate Research Center, San Jose, CA \\
    {\tt\small \{mx6, djcran\}@indiana.edu, chen1.xin@midea.com}
}

\maketitle
\ifwacvfinal\thispagestyle{empty}\fi

\begin{abstract}
    A major emerging challenge is how to
    protect people's privacy
    as cameras and computer vision are increasingly integrated into our daily lives,
    including in smart devices inside  homes.
    A potential solution is to capture and
    record just the minimum amount of information needed to perform a task
    of interest.  In this paper, we propose a fully-coupled two-stream
    spatiotemporal architecture for reliable human action
    recognition on extremely low resolution (e.g., 12$\times$16 pixel)
    videos. We provide an efficient method to extract spatial and temporal
    features and to aggregate them into a robust feature representation
    for an entire action video sequence. We also consider how to
    incorporate high resolution videos during training in order to
    build better low resolution action recognition models.  We evaluate on
    two publicly-available datasets, showing significant improvements over
    the state-of-the-art.
\end{abstract}


\section{Introduction}

Cameras are seemingly everywhere, from the traffic cameras in cities
and highways to the surveillance systems in businesses and public
places. Increasingly we allow cameras even into the most private
spaces in our lives: gaming consoles like Microsoft
Kinect~\cite{kinect} watch our living rooms, ``smart home'' devices
like Amazon Echo Look~\cite{echolook} and Nest Cam~\cite{nestcam}
monitor our homes, ``smart toys'' like the Fisher-Price Smart Toy
Bear~\cite{toybear} entertain our children, ``smart appliances'' like
the Samsung Family Hub~\cite{familyhub} watch and respond to our
everyday actions, and cameras in mobile devices like smartphones and
tablets see us even in our bedrooms. While these cameras have the
promise of making our lives safer and simpler, including making
possible more natural, context-aware interactions with technology,
they also record highly sensitive information about people and their
private environments.

\begin{figure}[t]
    {{
            \begin{minipage}[b]{1.0\linewidth}
                \centerline{\includegraphics[height=2.5cm, width=8.3cm]{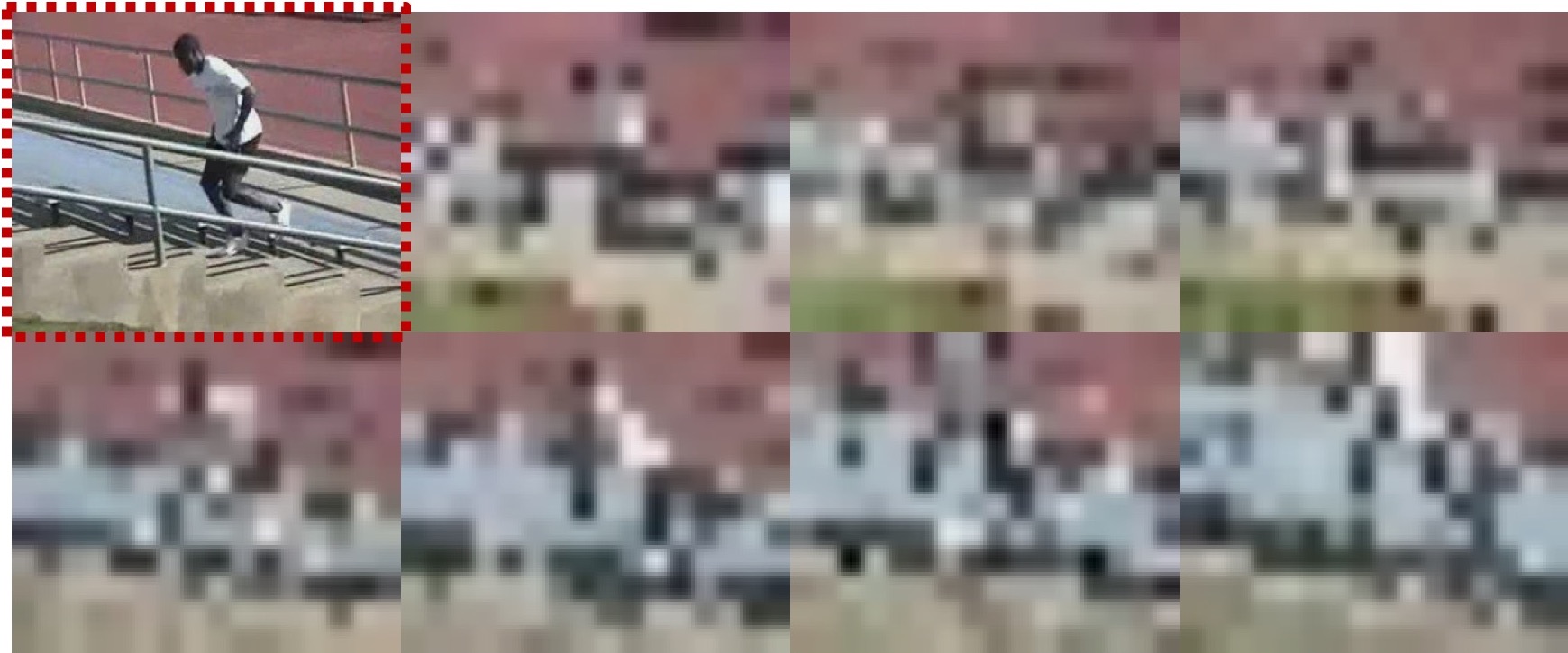}}
                \centerline{\textcolor{black}{(a) Climb Stairs}}\smallskip
            \end{minipage}
            \begin{minipage}[b]{1.0\linewidth}
                \centerline{\includegraphics[height=2.5cm, width=8.3cm]{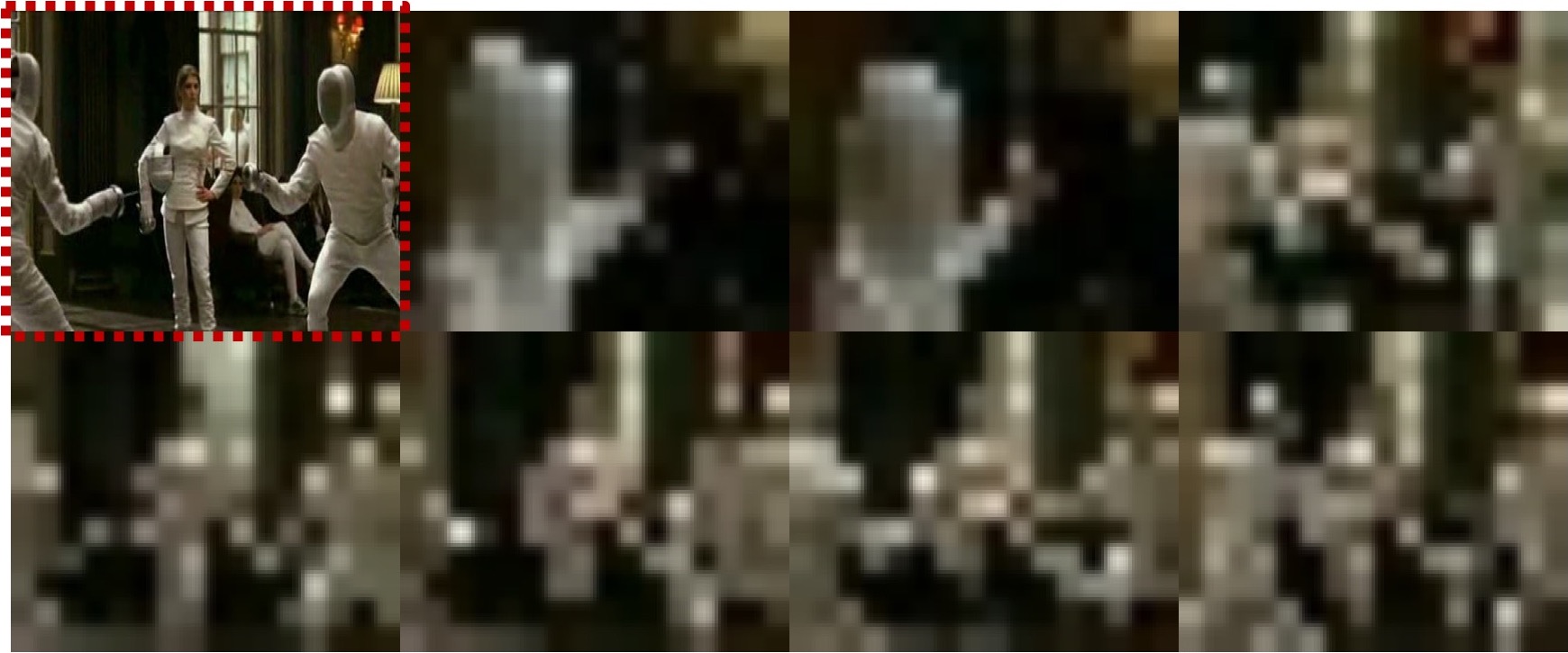}}
                \centerline{\textcolor{black}{(b) Fencing}}\smallskip
            \end{minipage}
            \begin{minipage}[b]{1.0\linewidth}
                \centerline{\includegraphics[height=2.5cm, width=8.3cm]{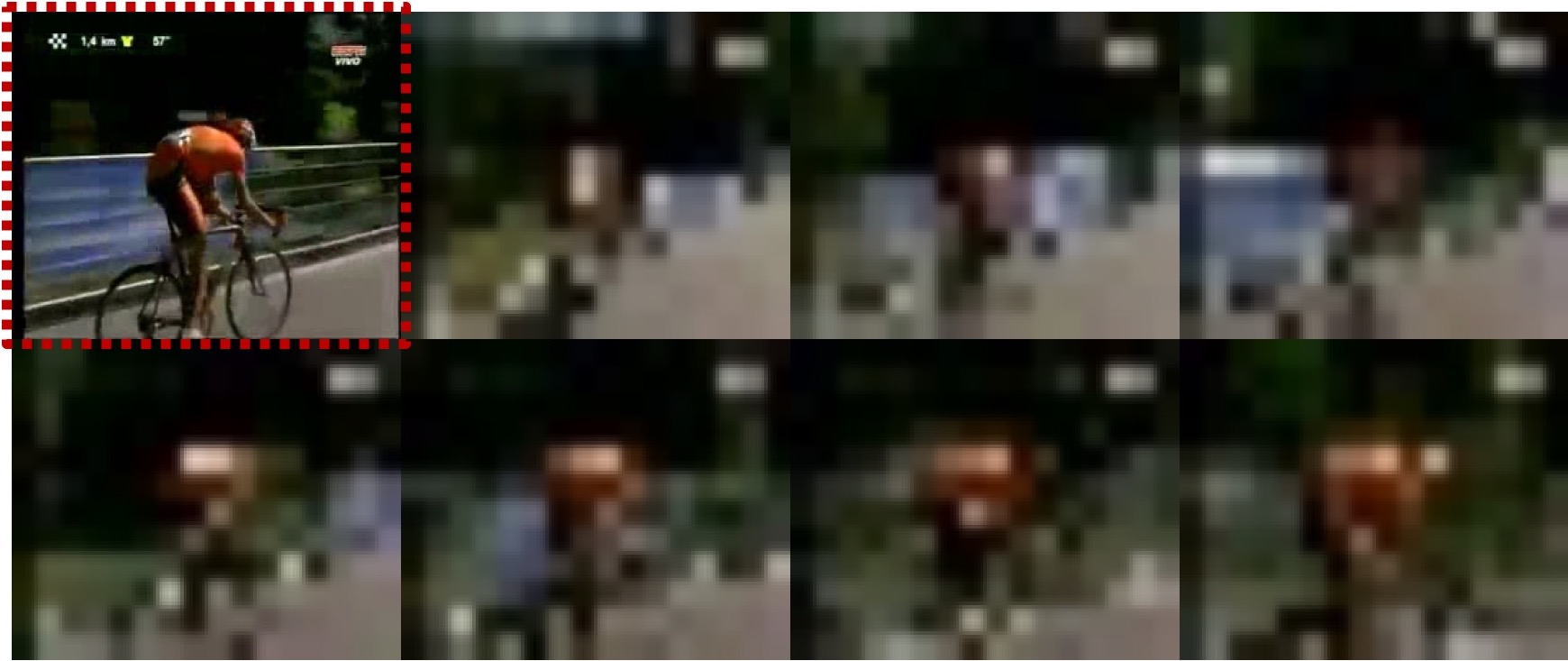}}
                \centerline{\textcolor{black}{(c) Ride Bike}}\smallskip
            \end{minipage}
    }}
    \vspace{-5pt}
    \caption{
        Sample frames of extremely low resolution ($12\times16$ pixel) videos
        streams from the HMDB51 dataset. 
        Original high resolution frames are shown in red.
    }
    \vspace{-5pt}
    \label{fig:teaser}
\end{figure}

To make matters worse,
processing for many of today's devices is often performed by remote
servers ``in the cloud.'' This means that even if a user trusts that a
device is using recorded video solely for legitimate
purposes (which is already a leap of faith, given cases of private
data being used for marketing and other unscrupulous
purposes~\cite{cnbc}), and that the device has been adequately
protected against hacking (another leap of
faith~\cite{placeraider,nsacamerahack,smarttoys}), a user must also
trust a remote cloud provider and the interlying network that their data must traverse.
Even a secure and trusted cloud
provider may be forced to share data with a government by a court order
or intelligence agency request~\cite{amazonmurder}.

One way of addressing 
the privacy challenge
is to transmit just
the minimum amount of information needed for a computer vision task to
be accurately performed. In the security community, solutions based on
selective encryption of image content~\cite{puppies} and
firmware-enforced access control~\cite{scannerdarkly} have been
proposed to keep video data safer from hacks, but service providers
must still be trusted with high-fidelity video content. Another
strategy is to remove private details in imagery before they leave the device. While techniques like
selective image blurring~\cite{blurcollaboration},
obscuring~\cite{markit}, and various other
transformations~\cite{facemorph,visualprivacysurvey,cartooning} have
been studied~\cite{viewerexperience2018chi}, these all require accurately defining and detecting
private content, which is in itself a highly non-trivial problem
(and thus these techniques are often not effective in practice~\cite{blurfails}).

Perhaps the most effective approach is to simply avoid collecting
high-fidelity imagery to begin with.  For example, low resolution
imagery may prevent specific details of a scene from being identified
-- e.g.\ the appearance of particular people, or the identity of
particular objects -- while still preserving enough information for a
task like scene type recognition~\cite{torralba200880}.  Particularly
important in many home applications of cameras is action and activity
recognition, to help give smart devices high-level contextual
information about what is going on in the environment and how to react
and interact accordingly.  Several recent papers have shown that very
low resolution videos (around $16\times12$ pixels) preserve enough
information for fine-grained action
recognition~\cite{dai2015towards,ryoo2017privacy,chen2017semi,ryoo2017low}.
This is perhaps surprising, since even a human observer may have
difficulty identifying actions from such little information 
(Figure~\ref{fig:teaser}).
This raises the question of how much better action recognition on
low resolution frames can progress.

Existing work on low resolution action recognition tends to focus on
modeling the spatial (appearance) information in each individual
frame. For example, a common approach is to use high resolution
training videos to learn a transformation between high and
low resolution frames, to help recover lost visual
information~\cite{ryoo2017privacy}.  This has been implemented by either
semi-coupled networks sharing convolutional filters between high and
low resolution inputs~\cite{chen2017semi} or Multi-Siamese networks
learning inherent properties of low resolution~\cite{ryoo2017low}.
However, much (if not most) useful information about
action recognition in low resolution video is in the \textit{motion}
information, not in any single frame.  

In this paper, we propose a fully-coupled two-stream
spatiotemporal network architecture to better take advantage of both local
and global temporal features for action recognition in low resolution
video. Our architecture incorporates motion information at three
levels: (1) a two-stream network incorporates stacked optical flow images
to capture subtle spatial changes in low resolution videos; (2) a 3D Convolution
(C3D) network computes temporal features within local video intervals; (3)
a Recurrent Neural Network (RNN) uses the extracted C3D features from videos
and optical flow fields to model more robust longer-range features. Our experiments on two
challenging datasets (HMDB51 and DogCentric) show that our model
significantly outperforms the previous state-of-the-art.

\section{Related Work}

Most state-of-the-art techniques for action recognition in video use
deep learning methods.  At a very high level, there are two important
types of evidence about action: appearance (spatial) features within
individual frames, and motion (temporal) features that cue on
distinctive movement patterns.  Karpathy \textit{et
al.}~\cite{karpathy2014large} were among the first to study deep
learning-based action recognition, proposing a multiresolution CNN
architecture that operates on individual frames without explicitly
modeling temporal information.  Simonyan and Zisserman~\cite{simonyan2014two}
used a two-stream CNN framework to incorporate both feature types,
with one stream taking RGB image frames as input and the other taking
pre-computed stacked optical flows.  The additional stream
significantly improved action recognition accuracy, indicating the
importance of motion features.  Tran \textit{et
al.}~\cite{tran2014learning} avoided the need for precomputing
optical flow features through their 3D convolution (C3D) framework,
which allows deep networks to learn temporal features in an end-to-end
manner.

Diba \textit{et al.}~\cite{diba2016deep} combined these two ideas into
two-stream C3D networks, and also proposed a more robust fusion method
for better temporal information encoding.  Zhu \textit{et
al.}~\cite{zhu2017action} avoid pre-computing optical flow, instead
learning the motion features in an end-to-end framework with a hidden
two-stream network. That approach is about ten times faster than
having to pre-compute optical flow, but the accuracy is somewhat weaker.
Most of these papers capture motion information over relatively short
temporal intervals, although several recent papers generate action
proposals for longer videos with a combination of C3D and Recurrent
Neural Networks (RNNs)~\cite{escorcia2016daps, sst_buch_cvpr17,
gao2017turn}.

Several papers have focused on recognition in
extremely low resolution (LR) imagery.  This problem is considered more
difficult, of course, since there is simply less visual information
available at low resolutions~\cite{zou2012very}. Wang \textit{et
al.}~\cite{wang2016studying} address low resolution object recognition
by taking advantage of high resolution training images to learn a transformation
between the two resolutions.
Dai \textit{et al.}~\cite{dai2015towards} adapted this idea to action recognition in the
video domain, specifically focusing on extracting and learning better
low resolution features from limited information.  Ryoo \textit{et
al.}~\cite{ryoo2017privacy} defined low resolution to be around $16
\times 16$ to $12 \times 16$ pixels and decomposed high resolution training videos
into multiple LR training videos by learning different resolution
transformations.  Chen \textit{et al.}~\cite{chen2017semi} used
two-stream semi-coupled networks to design an end-to-end training
network on both visual and motion information. By observing that two
LR images taken from exactly the same scene can contain totally
different pixel values, recent follow-up work by Ryoo \textit{et
al.}~\cite{ryoo2017low}  achieved state-of-the-art LR action recognition
performance  by
learning an embedding representation with Multi-Siamese networks.

We build on these previous methods that focus mostly on modeling and encoding spatial
features from low resolution video frames, and propose an
action recognition approach that incorporates stronger motion information.
In particular, our model captures motion information within different
temporal neighborhoods, including both sequential dependencies between
consecutive frames and more global temporal features.
We find that this, combined with a fully-coupled network that learns
from high resolution training videos, yields stronger models that 
significantly outperform  state-of-the-art methods.

\begin{figure}
    \begin{minipage}[b]{1.0\linewidth}
        \centerline{\includegraphics[width=8.5cm]{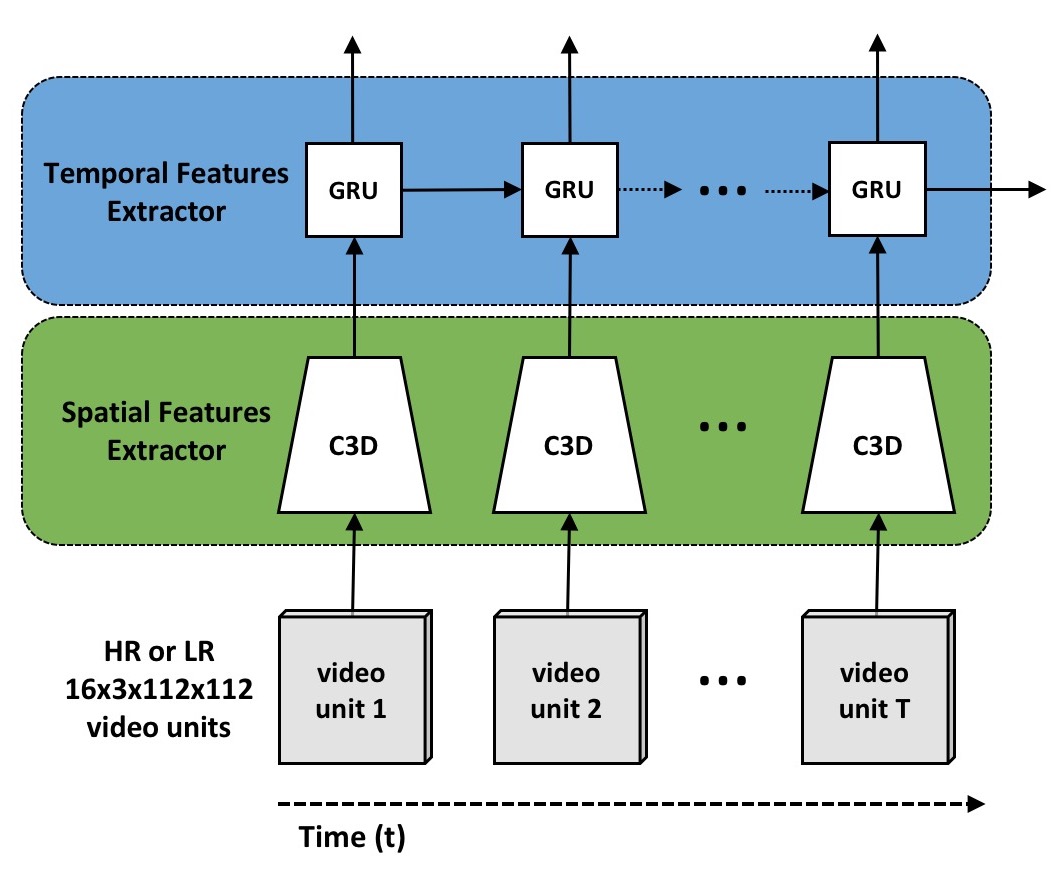}}
    \end{minipage}
    \vspace{-8pt}
    \caption{
        Visualization of our spatiotemporal features extractor, which
uses a C3D network to capture spatial and temporal
        features for video units and an RNN to encode  motion
        information across the entire video stream.
    }
    \label{fig:spatiotemporal}
\end{figure}

\section{Technical Approach}

\begin{figure*}
    \begin{center}
        \includegraphics[width=17.5cm]{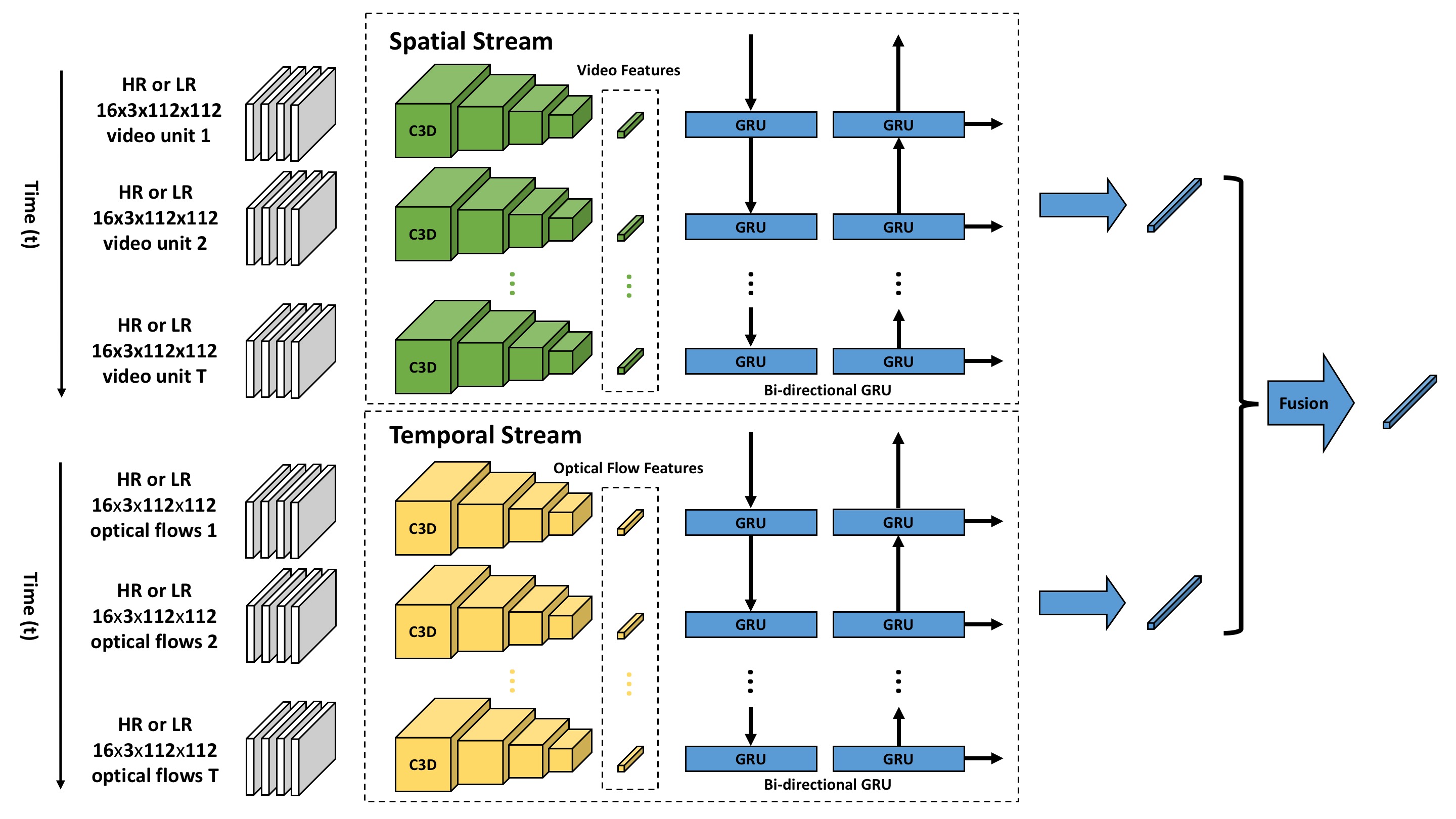}
    \end{center}
    \caption{ Visualization of our fully-coupled two-stream
      spatiotemporal networks. We feed RGB frames into the
      spatial stream (green) and corresponding stacked
      optical flow fields to the temporal stream (yellow). The
      GRU networks (blue) compute spatiotemporal
      information for the entire video using the extracted C3D
      features as inputs. In training, both C3D and GRU are
      fully-coupled with convolution filters shared between high and low resolution
      training videos, whereas only low resolution videos are used in testing.  }
\label{fig:overall}
\end{figure*}

We now present our approach for action recognition on extremely low
resolution videos. Specifically, we first introduce the basic
architecture of our spatiotemporal feature extractor, which uses a
combination of 3D Convolutional (C3D) Networks and Recurrent Neural
Networks (RNNs). Then, we discuss how to learn transferable features
from high to low resolution videos, which is based on the assumption
that high resolution \textit{training} videos are available. Finally,
we explore four fusion methods to efficiently combine visual
and motion information during recognition. 

\subsection{Spatiotemporal Feature Extractor}

We assume we are given an input video sequence
with $L$ frames that has captured a person or people performing a
single action (e.g., kicking a ball, shaking hands, chewing food,
etc.) and our goal is to recognize this action. Unlike prior work that
has viewed each frame as a separate processing unit, we discretize the
video into $T = L/\delta$ non-overlapping  \textit{video
units} $V_1, ... V_T$, each containing $\delta$ consecutive frames. Figure~\ref{fig:spatiotemporal}
illustrates the general architecture of the extractor network.

\xhdr{Spatial Feature Extractor.} Motivated by Buch \textit{et al.}'s
work~\cite{sst_buch_cvpr17} on high resolution video, we propose a
feature extractor to characterize appearance information in
low resolution video units. In particular, we use the C3D network,
which has proven to be well-suited for modeling sequential inputs such
as videos~\cite{tran2014learning}.  Since C3D uses 3D
convolution and pooling operations that operate over both spatial and
temporal dimensions, it is able to capture motion information within
each input video unit. We successively feed each video
unit $V_t$ to C3D, and extract its feature
representation $x_t$ from the last fully-connected layer. 

\xhdr{Temporal Feature Extractor.} While the C3D network is able to
encode local temporal features within each video unit, it cannot model
across the multiple units of a video sequence. We thus introduce a
Recurrent Neural Network (RNN) to capture global sequence dependencies
of the input video and cue on motion information (e.g., trajectories).
Although Long Short-Term Memory (LSTM)~\cite{hochreiter1997long}
networks are the most widely used RNNs in video
applications, we found that Gated Recurrent Units
(GRUs)~\cite{chung2015gated} performed better in our application. The
basic mechanism behind GRUs is similar to LSTMs, except that they do
not use memory units to control the information flow, and they have
fewer parameters which makes them slightly easier and faster to use
both in training and testing. 

More formally, the GRU is trained to take the extracted C3D features
$x_1, x_2, ..., x_T$ and output a sequence of hidden states $h_1, h_2,
\cdots, h_T$, with learnable hidden layer parameters $W$, $U$, and $b$.
Then, the GRU cell iterates the following operations
for $t \in [1, T]$:
\begin{equation}
    \begin{split}
        &z_t = \textrm{sigmoid} (W_z x_t + U_z h_{t-1} + b_z), \\
        &r_t = \textrm{sigmoid} (W_r x_t + U_r h_{t-1} + b_r), \\
        &n_t = \textrm{tanh} (W_h x_t + U_h (r_t \circ h_{t-1}) + b_h), \mbox{ and} \\
        &h_t = z_t \circ h_{t-1} + (1 - z_t) \circ n_t,
    \end{split}  
\end{equation}
where $\circ$ is the Hadamard product, and $z_t$, $r_t$, $n_t$, and $h_t$ are
the reset, input, new gates, and hidden state for time $t$, respectively.

The hidden state $h_t$ at each time $t$ is a feature vector
representing encoded spatiotemporal information corresponding to the
first $t$ video units, so $h_T$ incorporates features of the entire
video.  We use $h_T$ as input to a fully-connected layer with a
softmax classifier to output the confidence score of each action
class. 
We call the spatiotemporal feature extractor the
\emph{spatial stream} of our two-stream networks.

\subsection{Fully-coupled Networks}
Low resolution recognition approaches in both
image and video domains have achieved better performance by learning
transferable features from high to low resolutions.  This process can
be done either using unsupervised pre-training on super resolution
 sub-networks~\cite{wang2016studying} or with partially-coupled
networks~\cite{wang2016studying, chen2017semi} which are more flexible
for knowledge transformation. Inspired by these results, we propose a
fully-coupled network architecture where all parameters of both the C3D
and GRU networks are shared between high and low resolutions in the (single) training
stage. The key idea is that by viewing high and low resolution video frames as two
different domains, the fully-coupled network architecture is able to
extract features across them.  Since high resolution video contains much more
visual information, training on both resolutions helps improve learning
spatial features; using high resolution in training can be thought of as 
data augmentation, since different techniques for sub-sampling
produce different low resolution examplars from the same original high resolution image.

\subsection{Two-stream Networks}
In this subsection, we extend our single-stream networks to two-stream networks by
adding a similar architecture but with optical flow fields as the input. Since 
motion features between consecutive low resolution video frames are often quite  small,
our model benefits from optical flow images to learn 
pixel-level correspondences of temporal features. We call this
module the \emph{temporal stream}. Figure~\ref{fig:overall} shows the main
architecture of our two-stream networks.

In particular, we compute optical flow fields for both high and low resolution
videos, following Chen \textit{et al.}~\cite{chen2017semi} and using
the public MATLAB toolbox of Liu~\cite{liu2009beyond}.
The optical flows are encoded as 3-channel HSL images,
where hue and saturation are converted from optical flow vectors
(x- and y-components) into polar coordinates and lightness is set to one.
Before computing optical flow, we downsample the high resolution frames and upsample
the low resolution frames to a common size of $112 \times 112$ pixels.

We also explore four widely used fusion
methods, which enable the model to leverage joint visual and motion information
more effectively.
Following the notational convention of Feichtenhofer \etal \cite{feichtenhofer2016convolutional},
a fusion function
$f : \textbf{x}^a_t, \textbf{x}^b_t \to \textbf{y}_t$
fuses two feature maps
$\textbf{x}^a_t \in \mathbb{R}^{H \times W \times D}$
and
$\textbf{x}^b_t \in \mathbb{R}^{H' \times W' \times D'},$
at state $t$, to output a map
$\textbf{y}_t \in \mathbb{R}^{H'' \times W'' \times D''},$
where $W$, $H$, and $D$ denote the width, height and number of channels, respectively. Different
papers apply the fusion function to feature maps at different points in a deep network,
while our model only applies
it to the final hidden state $h_T$. Since the hidden state
of a GRU is a $D$-dimensional vector, for simplicity, we drop the width, height
and $t$ subscripts and assume that $D = D' = D''$.

\begin{figure*}
    \begin{center}
{{
{\hspace{1cm} \textbf{HMDB51}} \hspace{6.8cm} {\textbf{DogCentric}} \\[1ex]
\begin{minipage}{12pt} 
\rotatebox{90}{
{\textbf{Optical flow fields \hspace{1.3in} 
Images \hspace{0.25in}}}}
\end{minipage}
\begin{minipage}{12pt} 
\rotatebox{90}{
{{Low resolution  \hspace{0.7cm}
High resolution  \hspace{0.5cm}
Low resolution  \hspace{0.7cm}
High resolution }}
}
\end{minipage}
\begin{minipage}{0.93\textwidth}
        \includegraphics[width=1\textwidth]{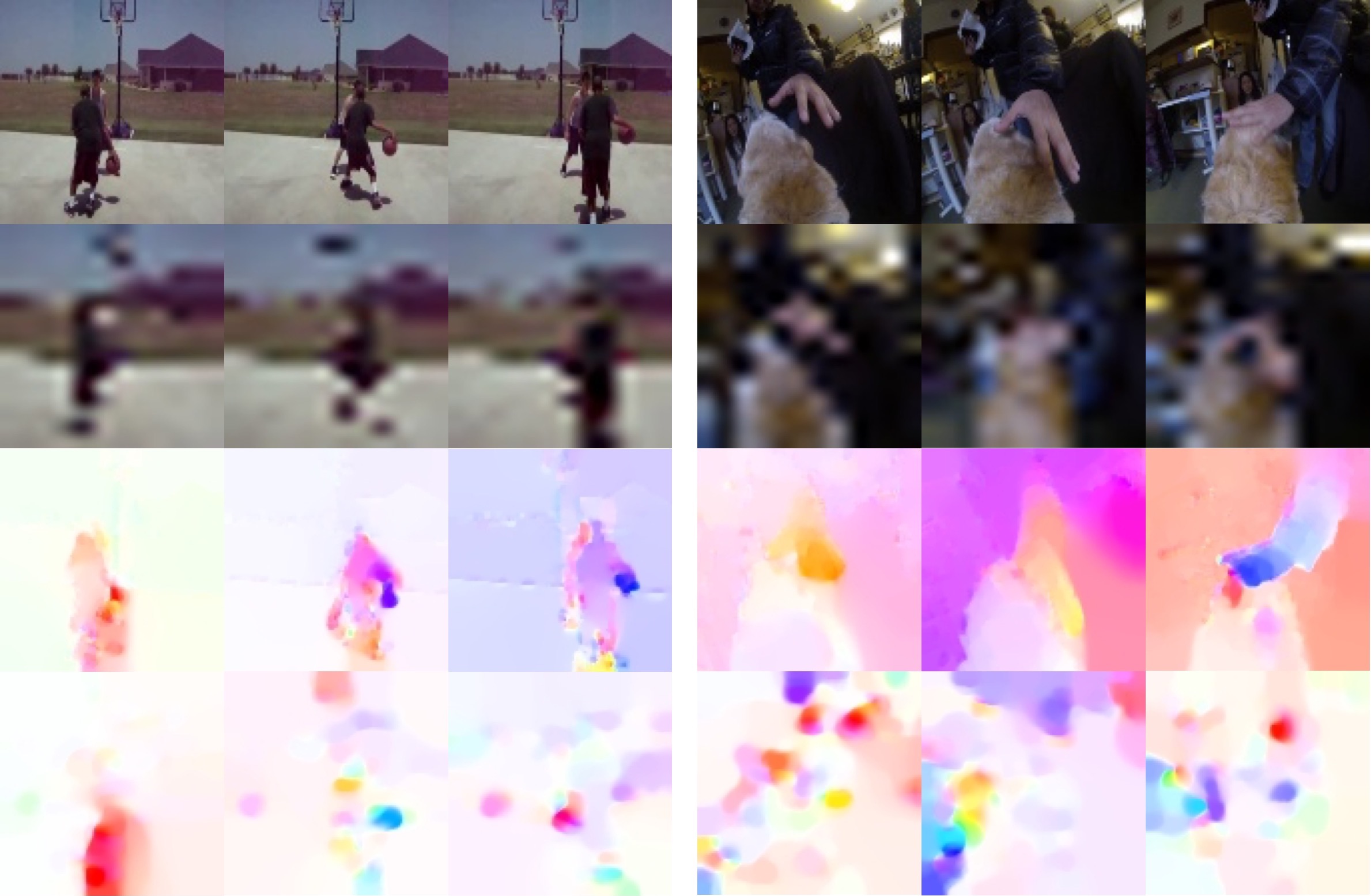}
\end{minipage}
}}
    \end{center}
    \caption{
        Examples of video frames from the HMDB51 (left) and DogCentric (right) datasets.
        \textit{First row:}
high resolution images 
        resized to 112$\times$112 pixels;
\textit{Second row:}
        low resolution  (12$\times$16 pixel) images upsampled to 
        112$\times$112 with bi-cubic interpolation; \textit{Third and fourth rows:} 
        optical flow fields for high resolution and low resolution
        images, respectively, calculated from images 
        rescaled to 112$\times$112.
    }
\label{fig:samples}
\end{figure*}

We consider four different fusion techniques:

\begin{enumerate}
\item \textbf{Sum Fusion,}
$\textbf{y}^{\textrm{sum}} = f^{\textrm{sum}}(\textbf{x}^a, \textbf{x}^b),$
computes the sum of two feature vectors over each channel $d$,
\begin{equation} \label{eq:1}
    y^{\textrm{sum}}_d = x^a_d + x^b_d,
\end{equation}
where $1<d<D$ and $\textbf{x}^a, \textbf{x}^b, \textbf{y} \in \mathbb{R}^{D}$.
Sum fusion is based on the assumption that  dimensions of the two output vectors of a two-stream
network correspond to one another.

\item \textbf{Max fusion,}
$\textbf{y}^{\textrm{max}} = f^{\textrm{max}}(\textbf{x}^a, \textbf{x}^b)$,
computes the maximum of two feature vectors over each channel $d$,
\begin{equation} \label{eq:2}
    y^{\textrm{max}}_d = \max\{x^a_d, x^b_d\}.
\end{equation}
In contrast to sum fusion,
max fusion only keeps the feature with the higher response, but again assumes
that corresponding dimensions are comparable.

\item \textbf{Concatenation Fusion,}
$\textbf{y}^{\textrm{cat}} = f^{\textrm{cat}}(\textbf{x}^a, \textbf{x}^b),$
stacks two feature vectors along the channel dimension,
\begin{equation} \label{eq:3}
    y^{\textrm{cat}}_{2d} = x^a_d \quad \textrm{and} \quad
    y^{\textrm{cat}}_{2d-1} = x^b_d,
\end{equation}
where $\textbf{y} \in \mathbb{R}^{2D}$. Instead of defining the correspondence
between two feature vectors by hand, concatenation fusion leaves this to be learned by
the subsequent convolution layers.

\item Finally, \textbf{Convolution Fusion,} $\textbf{y}^{\textrm{conv}} = f^{\textrm{conv}}(\textbf{x}^a, \textbf{x}^b)$,
is similar to
concatenation fusion, but is convolved with a learnable bank
of filters
$\textbf{f} \in \mathbb{R}^{2D \times D^o}$ and biases $b \in \mathbb{R}^{D^o}$
are appended after the concatenation fusion,
\begin{equation} \label{eq:4}
    y^{\textrm{conv}} = y^{\textrm{cat}} * \textbf{f} + b,
\end{equation}
where $*$ denotes convolution and $D^o$ denotes the number of output channels.
Filter $\textbf{f}$ provides a flexible way for the fusion
method to measure the weighted combination of $\textbf{x}^a$ and
$\textbf{x}^b$, and is able to project the channel dimension from $2D$ to
$D^o$. In order to permit a fair comparison with other methods, we set
$D^o = D$ and $D^o = 2D$ in our experiments.

\end{enumerate}

\section{Experiments}

We now evaluate and report results of each of the above methods on
our problem of action recognition in very low resolution video sequences.

\subsection{Datasets}
We evaluated our proposed techniques on low resolution versions of the
HMDB51~\cite{kuehne2011hmdb} and DogCentric~\cite{yumi2014first}
benchmarks, which are among the most widely used datasets in action
recognition at extremely low resolutions. The HMDB51 (Human Motion
Database) dataset consists of 7,000 video clips from a variety of
sources ranging from movies to YouTube videos, and is annotated with
51 action classes such as eating, smiling, clapping, bike riding,
shaking hands, etc.  Each clip is approximately 2-3 seconds long and
is recorded at 30 frames per second.  We follow prior work and report
accuracies averaged over three splits of training and testing
data. The DogCentric dataset is collected from wearable cameras
mounted on dogs, and consists of about 200 videos categorized into 10
different actions (which include activities performed by the dog
itself (e.g., drinking, walking, etc.) and interactions between the
dog and people. In contrast to HMDB51 which captures actions from a
third-person perspective, these videos capture actions of the camera
wearer from a first-person viewpoint.

\xhdr{Low Resolution Videos.}  The above two datasets were recorded at
a resolution of $240\times320$ pixels.  Since our C3D network expects
frames at a resolution of $112\times112$ pixels, we subsample the high
resolution images (used only during training) to this size.  To
generate low resolution training and testing data, we resize the
original videos to $12\times16$ using average downsampling, then
upscale them back to $112\times112$ resolution using bi-cubic
interpolation. We do not introduce any extra evidence into the
interpolation operation to ensure the $112\times112$ videos have
no more information than the $12\times16$ videos.  Figure~\ref{fig:samples} 
shows several corresponding low and high resolution frames 
as examples.

\subsection{Implementation Details}
Our learning process consists of two stages: (1) training the C3D network
and extracting features for video units; and (2) training our fully-coupled two-stream
networks with the concatenated C3D features of each video as inputs.

For the C3D networks, we followed Tran \textit{et al.}~\cite{tran2014learning}
and used their publicly available
pre-trained model and fine-tuned on the HMDB51 and DogCentric datasets
in both high and low resolution. Since we also extracted C3D features for optical
flow inputs, we fine-tuned another C3D model with high and low resolution optical
flows. The network architecture had 8 convolution layers, 5
max-pooling layers, and 2 fully-connected layers.  The length of each
video unit was set to 16 frames.  We used the output of the first
fully-connected layer \emph{fc6} (which had 4096 dimensions) and
stacked them together to form the video descriptor.

We implemented the fully-coupled two-stream networks in PyTorch~\cite{pytorch}
and used the C3D features  and optical flow as inputs. As
discussed before, we used mixed (high and low resolution) data in the training stage,
but no high resolution information is used in testing. We used the root-mean-square
propagation (RMSprop)~\cite{hinton2012neural} update rule to learn the
network parameters with fixed learning rate $10^{-3}$ and weight decay
$0.0005$. The whole training process stopped at 50 epochs, with the
batch size set to 256. All our experiments were
conducted on a system with a Nvidia Titan X Pascal graphics card.

\begin{table*}\centering
{{
    {
        \ra{1.2}
        \begin{tabular}{@{}cccccccccc@{}} \toprule
            \multicolumn{7}{c}{Network Architecture} & & & \\
            \cmidrule{1-7}
            Type & & C3D & & RNN & & Fusion & & \textbf{Accuracy} \\ \midrule
            \multirow{2}{*}{Temporal Network} & & pre-trained & & w/o GRU & & Sum Fusion      & & 21.24\% \\
            & & pre-trained & & uni-directional GRU & & Sum Fusion                           & & 24.11\% \\ \midrule
            \multirow{2}{*}{Spatial Network}  & & pre-trained & & w/o GRU & & Sum Fusion      & & 34.90\% \\
            & & pre-trained & & uni-directional GRU & & Sum Fusion                           & & 39.15\% \\ \midrule
            \multirow{7}{*}{Two-stream Network} & & pre-trained    & & w/o GRU & & Sum Fusion & & 38.56\% \\
            & & pre-trained    & & uni-directional GRU & & Sum Fusion                        & & 43.38\% \\
            & & w/o pre-trained & & bi-directional GRU  & & Sum Fusion                        & & \textbf{41.04}\% \\
            & & pre-trained    & & bi-directional GRU  & & Sum Fusion                        & & \textbf{44.96}\% \\
            & & pre-trained    & & bi-directional GRU  & & Max Fusion                        & & 42.02\% \\
            & & pre-trained    & & bi-directional GRU  & & Concatenate Fusion                & & 43.13\% \\
            & & pre-trained    & & bi-directional GRU  & & Convolution Fusion                & & 43.46\% \\
            \bottomrule
        \end{tabular}
    }
}}
    \vspace{8pt}
    \caption{Evaluation results of each component of our network architecture
    on the HMDB51 dataset.}
    \label{tab:self}
\end{table*}

\subsection{Evaluation}
Table~\ref{tab:hmdb} shows results of the evaluation.
Our full model featuring
pre-trained C3D networks, the bi-directional GRU network, and
the fully-coupled two-stream architecture with sum fusion 
achieves $44.96\%$ accuracy on the low resolution  HMDB51 dataset and $73.19\%$ on the low resolution DogCentric dataset.
Of course, these results are significantly worse than the best results on the high resolution versions of these
datasets (e.g. around $80.7\%$ for HMDB51~\cite{carreira2017quo}).
(We also tested
our best model on action recognition on high resolution videos, and easily achieved over
$68\%$ accuracy without any explicit tuning on network architecture and hyperparameters.)
However, as shown in Tables~\ref{tab:hmdb} and~\ref{tab:dog}, our results do beat all state-of-the-art approaches on low resolution video,
including Pooled Time Series (PoT) (which uses a combination of HOG, HOF, CNN features)~\cite{ryoo2015pooled},
Inverse Super Resolution (ISR)~\cite{ryoo2017privacy},
Semi-coupled Two-stream Fusion ConvNets~\cite{chen2017semi},
and Multi-Siamese Embedding CNNs~\cite{ryoo2017low}.
Our best result outperforms these methods by 7.2\% (3.3\% without pre-training) on the HMDB51 dataset
and 3.7\% on the DogCentric dataset.

\begin{figure}
    \begin{center}
        \includegraphics[width=8.5cm]{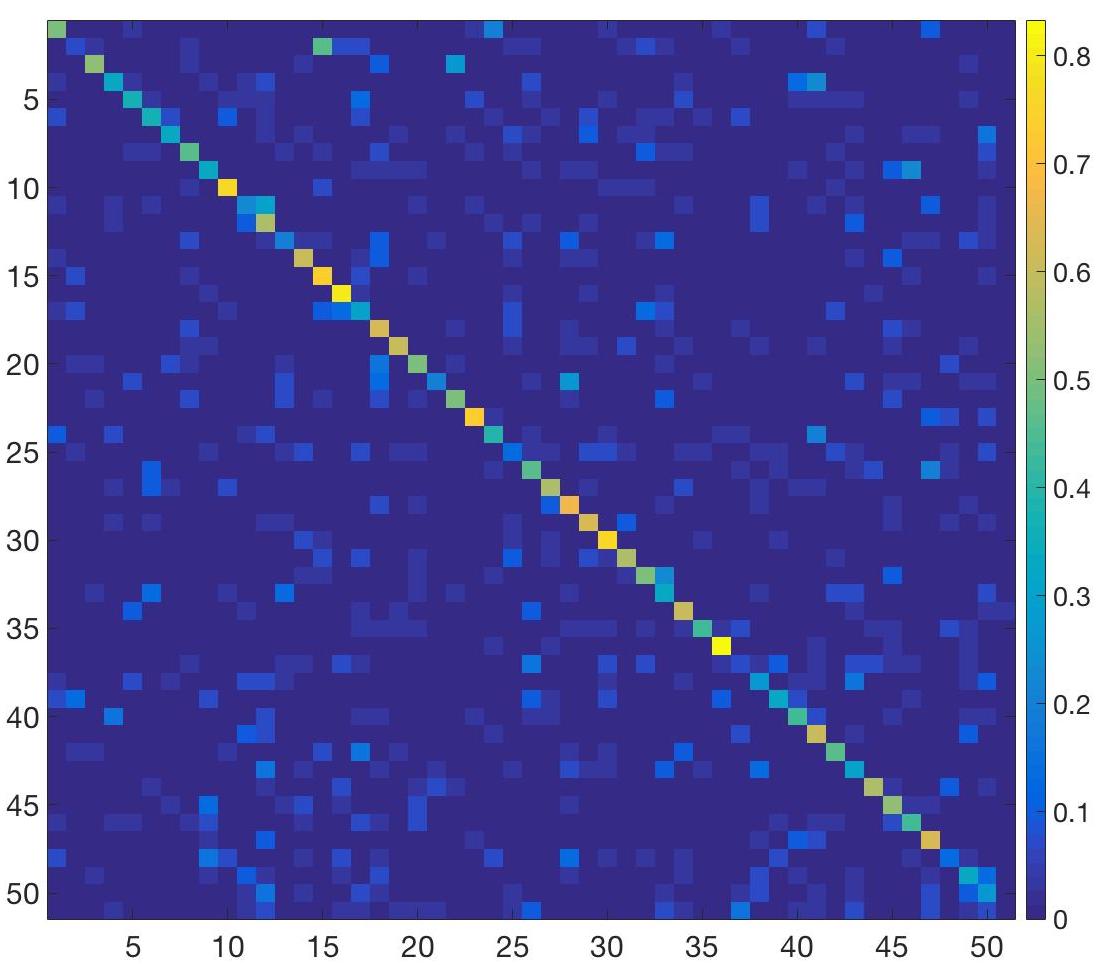}
    \end{center}
    \caption{
        Confusion matrix on the HMDB51 dataset using our best model.
        The x-axis denotes the predicted labels and the y-axis represents
        the ground truth labels for 51 action classes.
    }
\label{fig:confusion}
\end{figure}

To see how well our model performs on different categories, we also present
a confusion matrix in Figure~\ref{fig:confusion}. 

\subsection{Discussion}

To evaluate the contribution of each component of our model, we
also implemented multiple simpler baselines.

\vspace{6pt}
\noindent
\textbf{C3D Pre-training.}
To more fairly compare our model with that of Chen \etal~\cite{chen2017semi}, we experimented with training
the C3D networks from scratch on HMDB51, using the same architecture
and hyperparameters as the pre-trained network. As shown in Tables~\ref{tab:self}
and~\ref{tab:hmdb}, although
the result is about $3.9\%$ worse without pre-training on the Sport-1M dataset,
our model still achieves state-of-the-art results.

\vspace{6pt}
\noindent
\textbf{GRU Networks.}  To measure the contribution of the GRU
networks to our overall approach, we tried replacing them
with two fully-connected layers. To reduce the interference from the
two-stream networks, we also implemented two one-stream networks each
with the spatial and temporal features as inputs, and compared the
results between the ones with and without the GRU, respectively. The
results are presented in Table~\ref{tab:self}. It is clear that the models with
GRU outperform by about $4.5\%$. 
We also tested uni- and
bi-directional GRU architectures, and found that bi-directional GRUs perform
slightly better, as shown in Table~\ref{tab:self}.

\vspace{6pt}
\noindent
\textbf{Two-stream Networks.}
After evaluating the GRU networks, we now turn to the two-stream architecture,
where we believe that the pixel-level motion information acquired from optical
flow  can improve the model's ability in temporal feature encoding. As
shown in Table~\ref{tab:self}, it is clear that the two-stream
networks  significantly outperform 
one-stream networks with
both spatial ($4.2\%$ better) and temporal features ($19.2\%$ better), respectively. 
We also assessed the effect of our four
different fusion methods: (1) sum fusion, (2) max fusion, (3) concatenate
fusion, and (4) convolution fusion, as discussed in the previous section. The
results summarized in Table~\ref{tab:self} show that sum fusion achieves the best
performance.

\begin{table}\centering
{{
    {
        \ra{1.2}
        \begin{tabular}{@{}lccc@{}} \toprule
            \textbf{Approach} & \textbf{Accuracy} \\ \midrule
            3-layer CNN \cite{ryoo2017privacy} & 20.81\% \\
            ResNet-31 \cite{he2016deep} & 22.37\% \\
            PoT (HOG + HOF + CNN) \cite{ryoo2015pooled} & 26.57\% \\
            ISR \cite{ryoo2017privacy} & 28.68\% \\
            Semi-coupled Two-stream ConvNets \cite{chen2017semi} & 29.20\% \\
            Multi-Siamese Embedding CNN \cite{ryoo2017low} & 37.70\% \\ \midrule
            \textbf{Ours} (w/o pre-trained C3D) & \textbf{41.04}\% \\
            \textbf{Ours} (w/ pre-trained C3D) & \textbf{44.96}\% \\
            \bottomrule
        \end{tabular}
    }
}}
    \caption{Performance of our model compared to the state-of-the-art results on
    the HMDB51 dataset.}
    \label{tab:hmdb}
\end{table}

\begin{table}\centering
{{
    {
        \ra{1.2}
        \begin{tabular}{@{}lccc@{}} \toprule
            \textbf{Approach} & & \textbf{Accuracy} \\ \midrule
            PoT (HOG + HOF + CNN) \cite{ryoo2015pooled} & & 64.60\% \\
            ISR \cite{ryoo2017privacy} & & 67.36\% \\
            Multi-Siamese Embedding CNN \cite{ryoo2017low} & & 69.43\% \\ \midrule
            \textbf{Ours} (w/ pre-trained C3D) & & \textbf{73.19}\% \\
            \bottomrule
        \end{tabular}
    }
}}
    \vspace{7pt}
    \caption{Performance of our model compared to the state-of-the-art results on
    the DogCentric dataset.}
    \vspace{-10pt}
    \label{tab:dog}
\end{table}

\section{Conclusion}
We presented a new Convolutional Neural Network framework for action
recognition on extremely low resolution videos. We achieved
state-of-the-art results on the HMDB51 and DogCentric datasets with a
combination of four important components: (1) a fully-coupled network
architecture to leverage high resolution images in training in order to learn a
cross-domain transformation between low and high resolution feature
spaces; (2) 3D convolutional components which extract compact and
efficient spatiotemporal features for short video units; (3) a
Recurrent Neural Network (RNN) which considers long-range temporal
motion information; and (4) two network streams having both image
frames and stacked dense optical flow fields as input, in order to take into
account detailed motion features between adjacent video frames. We
hope this paper inspires more work on extremely low resolution action
recognition, and in methods to learn spatial-temporal features through
sequential images more generally.

\section{Acknowledgments}

This work was supported by the Midea Corporate Research Center
University Program, the National Science Foundation under grants
CNS-1408730 and CAREER IIS-1253549, and the IU Office of the Vice Provost for
Research, the College of Arts and Sciences, and the School of
Informatics, Computing, and Engineering through the Emerging Areas of
Research Project ``Learning: Brains, Machines, and Children.''
We thank Katherine Spoon, as well
as the anonymous reviewers, for helpful comments and suggestions on
our paper drafts.

{\small
    \bibliographystyle{ieee}
    \bibliography{egbib}
}

\end{document}